\documentclass{article}
\usepackage{graphicx}
\usepackage{amsmath}
\usepackage{amsthm}
\usepackage{amssymb}
\usepackage{algorithm, algpseudocode}
\usepackage{subfig}

\begin{document}

\title{\large \textbf{Unifying Model Predictive Path Integral Control, Reinforcement Learning, and Diffusion Models for Optimal Control and Planning}}

\author{
    \begin{minipage}[t]{0.45\textwidth}
        \centering
        \small
        \textbf{Yankai Li}\\
        School of Computing Science\\
        Simon Fraser University\\
        \texttt{yla890@sfu.ca}
        \vfill
    \end{minipage}
    \hfill
    \begin{minipage}[t]{0.45\textwidth}
        \centering
        \small
        \textbf{Mo Chen}\\
        School of Computing Science\\
        Simon Fraser University\\
        \texttt{mochen@cs.sfu.ca}
        \vfill
    \end{minipage}
}

\date{}

\maketitle

\begin{abstract}
Model Predictive Path Integral (MPPI) control, Reinforcement Learning (RL), and Diffusion Models have each demonstrated strong performance in trajectory optimization, decision-making, and motion planning. However, these approaches have traditionally been treated as distinct methodologies with separate optimization frameworks. In this work, we establish a unified perspective that connects MPPI, RL, and Diffusion Models through gradient-based optimization on the Gibbs measure. We first show that MPPI can be interpreted as performing gradient ascent on a smoothed energy function. We then demonstrate that Policy Gradient methods reduce to MPPI by applying an exponential transformation to the objective function. Additionally, we establish that the reverse sampling process in diffusion models follows the same update rule as MPPI.
\end{abstract}

\section{Preliminaries}

\subsection{Zeroth-order Optimization}

Zeroth-order (ZO) optimization, in contrast to its first-order (FO) counterpart, does not require explicit gradient information. Instead, it optimizes an objective function solely based on function evaluations, making it particularly useful in scenarios where gradient computations are unavailable. Several well-known optimization frameworks fall within the scope of ZO optimization, including: Evolutionary Optimization (e.g., Genetic Algorithms) \cite{goldberg1994genetic} and Bayesian Optimization (e.g., Gaussian Processes) \cite{shahriari2015taking}.

In the domain of control, Model Predictive Path Integral Control (MPPI) \cite{williams2016aggressive} is widely used for online motion planning due to its flexibility and ability to parallelize computations efficiently. The MPPI update rule is given by:

\begin{equation}
\label{eq:4}
U' = U + \frac{\sum^{N}_{i=1} \exp\left(- \frac{J(U+\mathcal{E}_i)}{\tau}\right)\mathcal{E}_i}{\sum^{N}_{i=1} \exp\left(- \frac{J(U+\mathcal{E}_i)}{\tau}\right)}
\end{equation}

where $J(\cdot)$ represents the cost function, $\tau$ is a temperature parameter, and $\mathcal{E}_i$ are sampled perturbations. We define 

\begin{equation}
U = (\mathbf{u}_0, \mathbf{u}_1, \cdots, \mathbf{u}_{T-1})\label{eq:22},
\end{equation}

\[\mathcal{E} = (\boldsymbol{\epsilon}_0, \boldsymbol{\epsilon}_1, \cdots, \boldsymbol{\epsilon}_{T-1}),\]

and

\[\boldsymbol{\epsilon} \sim \mathcal{N}(\mathbf{0}, \Sigma).\]

\cite{williams2017information} extends MPPI by incorporating a Bayesian prior on the control inputs, modeled as a Gaussian prior with zero mean. This formulation arises from the definition of control free energy $\mathcal{F}(V) = -\tau \log \left(\mathbb{E}_{U\sim\mathcal{N}(0, \Sigma)} \left[\exp\left(-\frac{1}{\tau} J(U)\right)\right]\right)$, leading to the modified MPPI update:

\[
U' = U + \frac{\sum^{N}_{i=1} \exp\left(- \frac{J(U+\mathcal{E}_i)}{\tau} - \sum_{t=0}^{T-1}\mathbf{u}_t^\top \Sigma^{-1}\boldsymbol{\epsilon}_{t,\,n}\right)\mathcal{E}_i}{\sum^{N}_{i=1} \exp\left(- \frac{J(U+\mathcal{E}_i)}{\tau} - \sum_{t=0}^{T-1}\mathbf{u}_t^\top \Sigma^{-1}\boldsymbol{\epsilon}_{t,\,n}\right)}
\]

where the additional summation term acts as a regularization term on $\mathbf{u}$. Specifically, if $\mathbf{u}$ and $\boldsymbol{\epsilon}$ share the same sign, the term penalizes it; conversely, if they have opposite signs, it encourages it.

\cite{williams2018information} further generalizes MPPI by allowing the base distribution to be any arbitrary distribution, rather than being restricted to Gaussians with zero mean. This results in the following generalized MPPI update:

\[
U' = U + \frac{\sum^{N}_{i=1} \exp\left(- \frac{J(U+\mathcal{E}_i)}{\tau} - \sum_{t=0}^{T-1}(\mathbf{u}_t - \widetilde{\mathbf{u}}_t)^\top \Sigma^{-1}\boldsymbol{\epsilon}_{t,\,n}\right)\mathcal{E}_i}{\sum^{N}_{i=1} \exp\left(- \frac{J(U+\mathcal{E}_i)}{\tau} - \sum_{t=0}^{T-1}(\mathbf{u}_t-\widetilde{\mathbf{u}}_t)^\top \Sigma^{-1}\boldsymbol{\epsilon}_{t,\,n}\right)}
\]

where $\widetilde{\mathbf{u}}_t$ represents the nominal control derived from the base distribution.

In this paper, we assume no control prior and use the MPPI formulation presented in Eq. \ref{eq:4}. We will show that this formulation is equivalent to performing gradient ascent on a smoothed energy function.

\bigskip

\begin{equation}
\label{eq:10}
P(\tau | \theta) = P(s_0)\prod_{t=0}^{T-1} P(s_{t+1}|s_t,a_t)\pi_\theta(a_t|s_t)
\end{equation}

\begin{equation}
\label{eq:9}
\underset{\tau \sim \pi_\theta}{E} [R(\tau)] = \int_\tau P(\tau | \theta) R(\tau) d\tau
\end{equation}

Similarly, Policy Gradient methods \cite{sutton1999policy} also belong to the category of zeroth-order (ZO) optimization, aiming to maximize the expected reward (Eq. \ref{eq:9}) by adjusting the parameter $\theta$. Like MPPI, they rely on sampling-based gradient estimation. However, a distinction is that Policy Gradient methods involve parameterized policies trained on datasets or rollouts, whereas MPPI operates in a non-parametric manner without the need for training. Policy Gradient has been instrumental in reinforcement learning (RL), with its variants, Trust Region Policy Optimization (TRPO) \cite{schulman2015trust} and Proximal Policy Optimization (PPO) \cite{schulman2017proximal}, playing significant roles in both control applications and large language models (LLMs).

\subsection{Diffusion}

Denoising diffusion models learn to map a simple prior distribution to a complex data distribution via a stochastic process. The data distribution is given by:

\[
\mathbf{x}_0 \sim q(\mathbf{x}_0)
\]

while the prior distribution is assumed to be a standard Gaussian:

\[
p(\mathbf{x}_T) = \mathcal{N}(\mathbf{x}_T;\;\mathbf{0},\mathbf{I})
\]

The forward process follows a Langevin equation \cite{lemons1997paul}, a stochastic differential equation (SDE) used to model diffusion dynamics:

\begin{equation}
\label{eq:prob_2}
d\mathbf{x} = -\nabla_{\mathbf{x}}\,U(\mathbf{x})dt + \sqrt{2}d \mathbf{w}
\end{equation}

where $U(\cdot)$ denotes a potential field.

Substituting the Gaussian prior distribution, the Langevin equation takes the following form:

\begin{align*}
d\mathbf{x} &= \nabla_{\mathbf{x}}\,\mathrm{log}\,\mathcal{N}(\mathbf{0}, \mathbf{I})dt + \sqrt{2}\;d\mathbf{w}\\
 &= -\mathbf{x} \; dt + \sqrt{2}\;d\mathbf{w}
\end{align*}

A more general formulation is:

\[
d\mathbf{x} = \mathbf{f}(\mathbf{x},t) dt + g(t) d\mathbf{w},
\]

where $\mathbf{f}(\mathbf{x}, t)$ represents the drift term, $\mathbf{w}$ denotes the Wiener process, and $g(t)d\mathbf{w}$ accounts for stochastic perturbations.

The reverse diffusion process, which is central to generative modeling, follows the reverse-time SDE:

\[
d\mathbf{x} = \left[\mathbf{f}(\mathbf{x},t) - g(t)^2 \nabla_{\mathbf{x}} \log p_t(\mathbf{x})\right] dt + g(t) d\bar{\mathbf{w}},
\]

where $\bar{\mathbf{w}}$ represents the backward Wiener process.

The training objective for learning a score function (i.e., the gradient of the log-density function) in diffusion-based models is given by:

\[
\underset{\theta}{\text{arg\,min}}\; \underset{p_{\text{data}}(\mathbf{x})}{\mathbb{E}}
\left[
\|\mathbf{s}_\theta(\mathbf{x}) - \nabla_{\mathbf{x}}\log\,p_{\text{data}} (\mathbf{x})\|^2_2
\right]
\]

This formulation can be rewritten as shown in \cite{song2019generative}, \cite{vincent2011connection}:

\[
\underset{\theta}{\text{arg\,min}}\; \underset{q(\mathbf{\widetilde{x}|x})p_{\text{data}}(\mathbf{x})}{\mathbb{E}}
\left[
\|\mathbf{s}_\theta(\mathbf{\widetilde{x}}) - \nabla_{\mathbf{\widetilde{x}}}\log\,q (\mathbf{\widetilde{x}}|\mathbf{x})\|^2_2
\right]
\]

This reformulation enables optimization at the level of individual data points, and the conditional probability $q(\mathbf{\widetilde{x}}|\mathbf{x})$ is straightforward to obtain as we will show in the Main section.

\subsection{Trajectory Optimization}

In this paper, we consider a general discrete-time dynamical system defined by the equation:

\begin{equation}
\label{eq:3}
    \mathbf{x}_{t+1} = \mathbf{F}(\mathbf{x}_t, \mathbf{u}_t)
\end{equation}

where $\mathbf{x}_t \in \mathbb{R}^n$ denotes the system state at time $t$, $\mathbf{u}_t \in \mathbb{R}^m$ is the control input at time $t$, and  $\mathbf{F}$ represents the (typically nonlinear) dynamics of the system, which may incorporate stochastic elements such as Gaussian noise.

The objective of trajectory optimization in this context is to find the optimal control sequence $U$ that minimizes the cost function

\[
J(U) = \sum^{T-1}_{t=0} C(\mathbf{x}_t, \mathbf{u}_t), + C_f(\mathbf{x}_{T})
\]

where $C(\mathbf{x}_t, \mathbf{u}_t)$ represents the running cost and $C_f(\mathbf{x}_{T})$ is the terminal cost.

\begin{equation}
\label{eq:1}
p(U, \mathbf{x}_0) = \frac{1}{Z(\mathbf{x}_0)} \exp(E(U, \mathbf{x}_0)/\tau)
\end{equation}

To solve this optimization problem, we employ gradient ascent on the \textbf{Gibbs measure} (Eq. \ref{eq:1}). We will show, in the Main section, that from this perspective, MPPI \cite{williams2017information}, \cite{williams2018information}, Policy Gradient \cite{sutton1999policy}, and Diffusion-based approaches \cite{janner2022planning} can all be understood as performing gradient ascent on this measure.

For clarity, we will omit dependence on $\mathbf{x}_0$ in subsequent expressions, as we focus on optimizing the control sequence under a fixed initial state.

We define the \textbf{energy function} according to the optimizaiton frameworks:

\begin{enumerate}
    \item For MPPI, energy corresponds to the negative of the cost function $J(U)$.
    \item For Policy Gradient, energy is defined as the accumulated rewards $R(U)$.
    \item For Diffusion-based models, energy is associated with the logarithm of the numerator in the data probability distribution $p(U) = \widetilde{p}(U) / Z$
\end{enumerate}

Formally, we define a unified energy representation as:

\[
E(U) := -J(U) := R(U) := \log \widetilde{p}(U)
\]

\section{Main}

Building on the unified energy representation outlined in the Preliminary section, we now establish connections between MPPI, Policy Gradient, and Diffusion-based approaches through the lens of gradient ascent on the Gibbs measure.

\subsection{MPPI}

The Gibbs free energy is defined as:

\begin{equation}
\label{eq:11}
G = E - \tau S
\end{equation}

In this equation,  $E$ represents the energy, while $\tau$ denotes the temperature, which serves as a scalar controlling the smoothness of the probability distribution. The term $S$ corresponds to entropy, preventing the probability distribution from collapsing to a Dirac delta function.

Unlike previous works \cite{williams2017information, williams2018information}, where MPPI was derived using control free energy:

\[
\mathcal{F}(V) = -\tau \log \left(\underset{U\sim\mathbb{P}}{\mathbb{E}} \left[\exp\left(-\frac{1}{\tau} J(U)\right)\right]\right)
\]

we define control free energy using the Gibbs free energy (Eq.\;\ref{eq:11}). A key distinction in our definition is that it does not require a base distribution $\mathbb{P}$, unlike \cite{williams2017information, williams2018information}. Then, the Gibbs measure (Eq.\;\ref{eq:1}) naturally provides the optimal control distribution, as it maximizes the Gibbs free energy:

\[
\underset{q}{\text{arg\,max}}\; \left(G_q\right) = \underset{q}{\text{arg\,max}}\; \left( \underset{U\sim q}{\mathbb{E}}[E(U)] + \tau S_q \right)  = p^\ast(U) = \frac{1}{Z} \exp(E(U)/\tau)
\]

where we denote $G_q$ as the Gibbs free energy corresponding to the probability distribution $q$, and $S_q$ as the entropy associated with $q$.

\begin{proof}

\begin{subequations}
\begin{align}
G_q &= \underset{U\sim q}{\mathbb{E}}[E(U)] + \tau S_q\\
&= \int E(U)q(U)dx - \tau \int q(U) \log q(U)dU\\
&= \int q(U) (E(U) - \tau\log q(U))dU\\
&= \tau \int q(U) \log \frac{\exp(E(U)/\tau)}{q(U)}dU\\
& \leq \tau \log \left( \int \exp(E(U)/\tau) dU \right)\label{eq:13}\\
& = \tau \log Z 
\end{align}
\end{subequations}

where at step (\ref{eq:13}) we applied Jensen’s inequality using the fact that the logarithm function is concave.

Next, we show that the Gibbs measure $p$ (Eq. \ref{eq:1}) maximizes $G$.

\begin{align*}
G_p &= \underset{U\sim p}{\mathbb{E}}[E(U)] + \tau S_p\\
&= \tau \int p(U) \log \frac{\exp(E(U)/\tau)}{p(U)}dU\\
&= \tau \int p(U) \log Z\, dU\\
& = \tau \log Z
\end{align*}

\end{proof}

To demonstrate the connection between MPPI updates and gradient ascent on a smoothed energy function, we define the target distribution as the convolution of the optimal control distribution with a Gaussian kernel $\phi \sim \mathcal{N}(0, \Sigma)$:

\begin{equation}
\label{eq:19}
q(U) := (p^\ast * \phi)(U)
\end{equation}

We refer to $q(U)$ as the Gaussian-smoothed optimal control distribution.
The energy function associated with this distribution is denoted as $\widetilde{E}(U)$:

\begin{equation}
\label{eq:16}
\widetilde{E}(U) := \log \left( \int \exp(E(y)) \phi(U - y) dy\right)
\end{equation}

which we call the \textbf{smoothed energy function}.

\begin{proof}

For simplicity, we set $\tau = 1$, though the same proof holds for any $\tau$.

\begin{align*}
q(U) &= \int \frac{\exp(E(y))}{Z} \phi(U - y) dy\\
&= \frac{1}{Z} \exp \left( \log \left( \int \exp(E(y)) \phi(U - y) dy\right) \right) \\
&= \frac{1}{Z} \exp(\widetilde{E}(U))
\end{align*}
\end{proof}

We now analyze the proximity of the smoothed energy function to the original energy function.

\begin{align*}
\widetilde{E}(U) &= \log \left( \int \exp(E(y)) \phi(U - y) dy\right)\\
&= \log \left( \underset{y\sim\phi(U-y)}{\mathbb{E}} [\exp(E(y))] \right)\\
&= \log \left( \underset{y\sim\phi(U-y)}{\mathbb{E}} [\exp(E(y) - E(U))] \right) + E(U)\\
\end{align*}

Then,

\[
\widetilde{E}(U) - E(U) = \log \left( \underset{y\sim\phi(U-y)}{\mathbb{E}} [\exp(E(y) - E(U))] \right)
\]

When the variance of $\phi$, denoted as $\Sigma$, is small, the distribution $y\sim\phi(U-y)$ forms a sharp Gaussian centered at $U$. Consequently, the expectation $\underset{y\sim\phi(U-y)}{\mathbb{E}} [\exp(E(y) - E(U))] \approx 1$, leading to

\[
\widetilde{E}(U) \approx E(U)
\]

By applying Jensen's inequality, a lower bound on $\widetilde{E}(U)$ can be derived:

\begin{align*}
\widetilde{E}(U) - E(U) &= \log \left( \underset{y\sim\phi(U-y)}{\mathbb{E}} [\exp(E(y) - E(U))] \right)\\
&\geq \underset{y\sim\phi(U-y)}{\mathbb{E}} \left[ E(y) - E(U) \right]\\
&= \int E(y) \phi(U - y) dy - E(U)
\end{align*}

Thus,

\[
\widetilde{E}(U) \geq \int E(y) \phi(U - y) dy
\]

which means that $\widetilde{E}(U)$ is at least the local Gaussian averaged $E(U)$.

We then show that MPPI can be rewritten as performing gradient ascent on the smoothed energy function. Applying gradient ascent, with learning rate $\left(\frac{1}{\tau}\right)\Sigma$, on this smoothed energy function results in the following update rule:

\begin{subequations}
\label{eq:15}
\begin{align}
U' &= U + \left(\frac{1}{\tau}\right)\Sigma \cdot \nabla \widetilde{E}(U)\\
&=U + \Sigma \cdot \nabla \left(\frac{\widetilde{E}(U)}{\tau}\right) - \Sigma\cdot\nabla \log \int \exp (\widetilde{E}(U)/\tau) dU   \label{eq:14}\\
& =U +  \Sigma \cdot \nabla\log \left( \frac{\exp(\widetilde{E}(U) / \tau)}{\int \exp(\widetilde{E}(U)/\tau) dU} \right)  \\
& =U +  \Sigma \cdot \nabla\log \left( \frac{\exp(\widetilde{E}(U) / \tau)}{Z} \right)  \\
&= U + \Sigma \cdot \nabla \log q(U)\\
&\approx U + \frac{\sum^{N}_{i=1} \exp\left(\frac{E(U+\mathcal{E}_i)}{\tau}\right)\mathcal{E}_i}{\sum^{N}_{i=1} \exp\left(\frac{E(U+\mathcal{E}_i)}{\tau}\right)} \label{eq:12}
\end{align}
\end{subequations}

At step (\ref{eq:14}), we used the fact that $Z=\int \exp (\widetilde{E}(U)/\tau) dU$ is a constant. A detailed proof of the step (\ref{eq:12}) can be found in \cite{xue2024full}.

The result in the derivation (Eq. \ref{eq:12}) aligns precisely with the MPPI update rule, establishing that MPPI is equivalent to performing gradient ascent on the smoothed energy function.

Additionally, our approach removes the assumption that the running cost must be decomposable into state cost and quadratic control cost, a requirement in \cite{williams2017information, williams2018information}. This makes our formulation applicable to a broader class of optimal control problems.

\subsection{Policy Gradient}

To establish the connection between Policy Gradient and MPPI, we consider a setting where the initial state in Policy Gradient is fixed for all trajectories, such that $s_0$ is deterministic. In this case, the initial state distribution $P(s_0)$ in Eq.\;\ref{eq:10} becomes a Dirac delta distribution centered at $s_0$, which is a common assumption in MPPI.

The original policy gradient formulation is given by:

\begin{subequations}
\begin{align}
\nabla_\theta \underset{\tau \sim \pi_\theta}{\mathbb{E}} [R(\tau)]
&= \nabla_\theta \int P(\tau | \theta) R(\tau)\,d\tau \\
&= \int \nabla_\theta P(\tau | \theta) R(\tau) \,d\tau \\
&= \int P(\tau | \theta) \nabla_\theta \text{log}\,P(\tau|\theta) R(\tau)\,d\tau \\
&= \underset{\tau \sim \pi_\theta}{\mathbb{E}} \left[\nabla_\theta \text{log}\,P(\tau|\theta)R(\tau)\right]\\
&= \underset{\tau \sim \pi_\theta}{\mathbb{E}} \left[
\sum_{t=0}^T \nabla_\theta \text{log}\,\pi_\theta(a_t|s_t)R(\tau)
\right] \label{eq:17}
\end{align}
\end{subequations}

For continuous action spaces, the policy $\pi_\theta(a|s)$ is typically modeled as a Gaussian distribution with mean $\mu_\theta$ and covariance $\Sigma_\theta$:

\begin{equation}
\label{eq:18}
\pi_\theta(a|s) = \mathcal{N}(\mu_\theta(s), \Sigma_\theta(s)) = (2\pi)^{-D/2} |\Sigma_\theta(s)|^{-1/2} \exp \left( -\frac{1}{2} (a-\mu_\theta(s))^\top \Sigma_\theta^{-1}(s) (a-\mu_\theta(s)) \right)
\end{equation}

Taking the logarithm of the policy distribution gives:

\[
\log \pi_\theta(a|s) = -\frac{D}{2} \log (2\pi) - \frac{1}{2} \log |\Sigma_\theta| - \frac{1}{2} (a-\mu_\theta)^\top \Sigma_\theta^{-1} (a-\mu_\theta)
\]

where $D$ is the dimensionality of $\mu$.

To align the formulation with MPPI, we assume that the variance $\Sigma$ is independent of $\theta$. Under this assumption, the expression simplifies to:

\[
\log \pi_\theta(a|s) = -\frac{D}{2} \log (2\pi) - \frac{1}{2} \log |\Sigma| - \frac{1}{2} (a-\mu_\theta)^\top \Sigma^{-1} (a-\mu_\theta)
\]

Note that the gradient of the log-policy with respect to $\theta$ can be rewritten as:

\[
\nabla_\theta \log \pi_\theta(a|s) = \nabla_{\mu_\theta} \log \pi_\theta(a|s) \cdot \nabla_\theta \mu_\theta
\]

Substituting this into Eq. \ref{eq:17}, we obtain:

\begin{equation}
\text{Eq.\;(\ref{eq:17})} 
= \underset{\tau \sim \pi_\theta}{\mathbb{E}} \left[
\sum_{t=0}^T \left(\nabla_{\mu_{t, \theta}} \log \pi_\theta(a_t|s_t) \cdot \nabla_\theta \mu_{t,\theta}\right) R(\tau)
\right]
\end{equation}

Since MPPI don't have parameters $\theta$, we drop the term $\nabla_\theta \mu_{t, \theta}$ and focus on $\nabla_{\mu_{t,\theta}} \log \pi_\theta(a_t|s_t)$ for the following discussion. Additionally, in MPPI, control actions at different time steps are independent, whereas in Policy Gradient, they are parameterized by $\theta$. Without loss of generality, we now consider an arbitrary time step $t$ and derive the expectation:

\begin{subequations}
\label{eq:23}
\begin{align}
&\underset{\tau \sim \pi_\theta}{\mathbb{E}} \left[ \left(\nabla_{\mu_{t, \theta}} \log \pi_\theta(a_t|s_t)\right) R(\tau)
\right]\\
=& \underset{\tau \sim \pi_\theta}{\mathbb{E}} \left[
 \left(\Sigma^{-1} (a_t-\mu_{t,\theta})\right) R(\tau)
\right] \label{eq:21}\\
=&\underset{\tau \sim \pi_\theta}{\mathbb{E}} \left[
 \left(\Sigma^{-1} \epsilon_t\right) R(\tau)
\right]\\
=& \Sigma^{-1} \underset{\tau \sim \pi_\theta}{\mathbb{E}} \left[
  \epsilon_t R(\tau)
\right]\\
=& \Sigma^{-1} \underset{\epsilon_t \sim \mathcal{N}(0, \Sigma)}{\mathbb{E}} \left[
  \epsilon_t R(\mathcal{E})
\right] \label{eq:20}\\
\approx& \Sigma^{-1} \frac{\sum^{N}_{i=1} \epsilon_{t,i} R(\mathcal{E}_i)}{N}
\end{align}
\end{subequations}

In steps (\ref{eq:21}) and (\ref{eq:20}) we used Eq. \ref{eq:18}\,.

The final step in reducing Policy Gradient to MPPI is introducing an exponential transformation to the objective function.

Instead of optimizing 

\[
\underset{\tau \sim \pi_\theta}{\mathbb{E}} [R(\tau)]
\]

we modify the objective to 

\[
\underset{\tau \sim \pi_\theta}{\mathbb{E}} [\exp (R(\tau))]
\]

which is equivalent to 

\[
Z \cdot\underset{\tau \sim \pi_\theta}{\mathbb{E}} [p(\tau)]
\]

where $p$ is the Gibbs measure (Eq. \ref{eq:1}).

With this objective, Eq. \ref{eq:23} becomes

\[
\underset{\tau \sim \pi_\theta}{\mathbb{E}} \left[ \left(\nabla_{\mu_{t, \theta}} \log \pi_\theta(a_t|s_t)\right) \exp(R(\tau))
\right] \approx \Sigma^{-1} \frac{\sum^{N}_{i=1} \epsilon_{t,i} \exp(R(\mathcal{E}_i))}{N}
\]

Extending this result to control actions across all time steps, we obtain:

\begin{align*}
&\underset{\tau \sim \pi_\theta}{\mathbb{E}} \left[ \left(\nabla_{\mu_{0, \theta}} \log \pi_\theta(a_0|s_0)\right) \exp(R(\tau)), \cdots, \left(\nabla_{\mu_{T-1, \theta}} \log \pi_\theta(a_{T-1}|s_{T-1})\right) \exp(R(\tau))\right]\\
= & \underset{\tau \sim \pi_\theta}{\mathbb{E}} \left[ \big(\nabla_{U}\big( \log \pi_\theta(a_0|s_0), \cdots, \log \pi_\theta(a_{T-1}|s_{T-1})\big)\big) \exp(R(U+\mathcal{E}_i))\right]\\
\approx & \Sigma^{-1}\frac{\sum^{N}_{i=1} \exp\left(R(U + \mathcal{E}_i)\right)\mathcal{E}_i}{N}
\end{align*}

Here we define $u_t := \mu_{t, \theta}$, and $U$ follows from Eq. (\ref{eq:22}).

Note that this formulation exactly matches MPPI (Eq.\;\ref{eq:4}) with the temperature term in MPPI accounting for the constant difference. As a result, the same derivation in Eq.\;\ref{eq:15} applies.

Thus, we find that both MPPI and Policy Gradient update perform gradient ascent on the smoothed energy function (Eq. \ref{eq:16}), provided that the objective for Policy Gradient is modified to $\underset{\tau \sim \pi_\theta}{\mathbb{E}} [\exp (R(\tau))]$ instead of the original $\underset{\tau \sim \pi_\theta}{\mathbb{E}} [R(\tau)]$.

\subsection{Diffusion}

The reverse process in diffusion models follows an update rule structurally identical to MPPI, which can be demonstrated through the discretized version of the reverse SDE \cite{song2021scorebased}:

\begin{equation}
\label{eq:8}
\mathbf{x}_i = \mathbf{x}_{i+1} - \mathbf{f}_{i+1} (x_{i+1}) + \mathbf{G}_{i+1} \mathbf{G}^\top_{i+1} \nabla_{\mathbf{x}_{i+1}} \log p_{i+1}(\mathbf{x}_{i+1}) + \mathbf{G}_{i+1} \mathbf{z}_{i+1}
\end{equation}

To maintain consistency with diffusion notations, we use $\mathbf{x}$ here; however, in our context, $\mathbf{x}$ corresponds to $U$.

For Variance Exploding (VE) Diffusion \cite{song2019generative}\cite{song2021scorebased}, the reverse update is given by:

\begin{subequations}
\begin{align*}
\mathbf{x}_i & \approx \mathbf{x}_{i+1} + (\sigma_{i+1}^2 - \sigma_i^2) \nabla_{\mathbf{x}_{i+1}} \log p_{i+1}(\mathbf{x}_{i+1}) + \sqrt{\frac{\sigma^2_{i}(\sigma^2_{i+1} - \sigma^2_{i})}{\sigma^2_{i+1}}}\mathbf{z}_{i+1} \quad \text{(Ancestral sampling)}\\
& \approx \mathbf{x}_{i+1} + (\sigma_{i+1}^2 - \sigma_i^2) \nabla_{\mathbf{x}_{i+1}} \log p_{i+1}(\mathbf{x}_{i+1}) + \sqrt{\sigma^2_{i+1} - \sigma^2_{i}}\mathbf{z}_{i+1} \quad \text{(Reverse diffusion samplers)}\\
& =  \mathbf{x}_{i+1} + (\sigma_{i+1}^2 - \sigma_i^2)  \nabla_{\mathbf{x}_{i+1}} \widetilde{E}(\mathbf{x}_{i+1}) + \sqrt{\sigma^2_{i+1} - \sigma^2_{i}}\mathbf{z}_{i+1}
\end{align*}
\end{subequations}

Similarly, for Variance Preserving (VP) Diffusion \cite{ho2020denoising} \cite{song2021scorebased}, the updates take the form:

\begin{subequations}
\begin{align*}
\mathbf{x}_i & \approx \frac{1}{\sqrt{1 - \beta_{i+1}}} \big(\mathbf{x}_{i+1} + \beta_{i+1} \nabla_{\mathbf{x}_{i+1}} \log p_{i+1}(\mathbf{x}_{i+1})\big) + \sqrt{\beta_{i+1}} \mathbf{z}_{i+1} \quad \text{(Ancestral sampling)} \\
& \approx \left( 2 - \sqrt{1 - \beta_{i+1}} \right) \mathbf{x}_{i+1} + \beta_{i+1} \nabla_{\mathbf{x}_{i+1}} \log p_{i+1}(\mathbf{x}_{i+1}) + \sqrt{\beta_{i+1}} \mathbf{z}_{i+1} \quad \text{(Reverse diffusion samplers)}\\
& \approx  \mathbf{x}_{i+1} + \beta_{i+1}  \nabla_{\mathbf{x}_{i+1}} \widetilde{E}(\mathbf{x}_{i+1}) + \sqrt{\beta_{i+1}} \mathbf{z}_{i+1}
\end{align*}
\end{subequations}

Note that ancestral sampling is derived from the ELBO, while reverse diffusion samplers is obtained from the discretized reverse process \cite{song2021scorebased}.

The final steps in the above derivations hold due to the definition of the marginal distribution:

\[
p_{i+1} (\mathbf{x}) = \int p_{i+1} (\mathbf{x}|\mathbf{x}_0)p_{\text{data}}(\mathbf{x}_0)d\mathbf{x}_0,
\]

where the conditional distribution for VE Diffusion follows:

\[
p_{i+1} (\mathbf{x}|\mathbf{x}_0) = \mathcal{N}(\mathbf{x};\; \mathbf{x}_0,\; \sigma^2\mathbf{I}),
\]

and for VP Diffusion, it is given by:

\[
p_{i+1} (\mathbf{x}|\mathbf{x}_0) = \mathcal{N}(\mathbf{x};\; \sqrt{\prod^i_{j=1} (1 - \beta_j)}\mathbf{x}_0,\; (1-\prod^i_{j=1} (1 - \beta_j))\mathbf{I}).
\]

Therefore, the reverse process in both VE and VP diffusion models follows an iterative update rule that closely aligns with MPPI.

\begin{figure}[h!]
    \centering
    \subfloat[Diffusion]{%
        \includegraphics[width=0.45\textwidth]{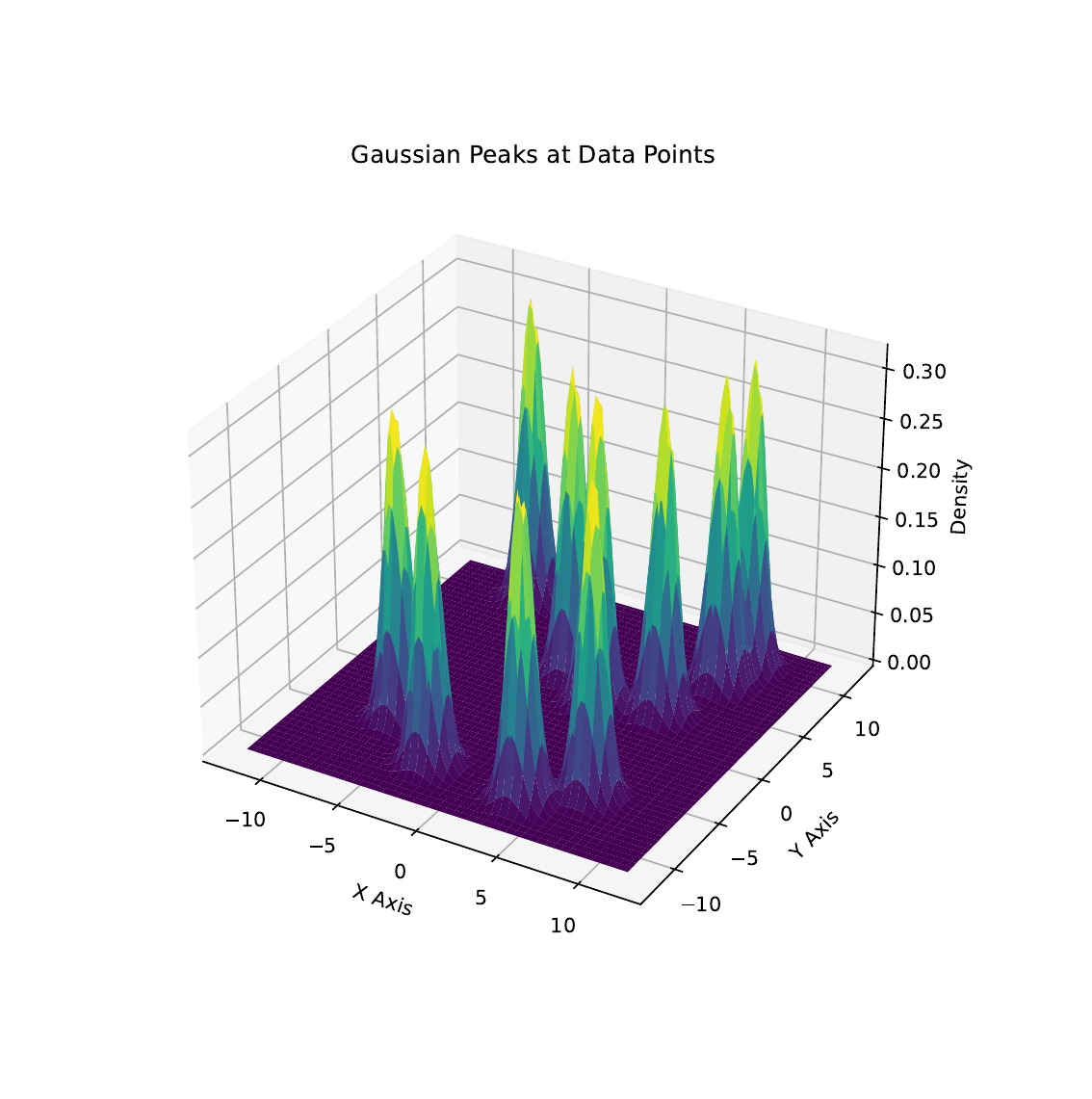}
        \label{fig:1}
    }
    \hfill
    \subfloat[MPPI and RL]{
        \includegraphics[width=0.45\textwidth]{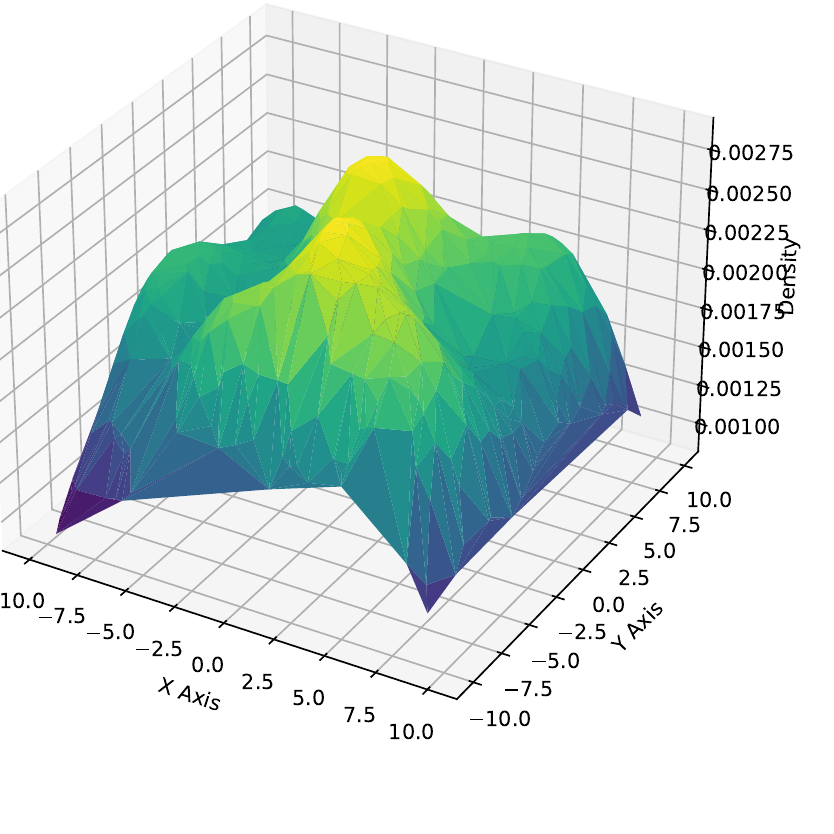}
        \label{fig:2}
    }
    \caption{Target Distribution}
\end{figure}

One difference between Diffusion models and MPPI/Policy Gradient lies in the nature of their optimization objectives. In diffusion-based control, the learning process is typically framed as supervised learning \cite{janner2022planning}, where the target distribution is directly shaped by the underlying data distribution, as illustrated in Figure \ref{fig:1}. In contrast, both MPPI and Policy Gradient determine the target distribution density based on cost (in MPPI) or reward (in Policy Gradient) as depicted in Figure \ref{fig:2}. This means that, unlike diffusion models, where the data-driven distribution determines sampling, MPPI and Policy Gradient derive their distributions directly from optimization objectives.

Guided diffusion-based planning also introduces a mechanism that aligns closely with MPPI. Guided diffusion-based planning has been actively explored in many studies, including Diffuser \cite{janner2022planning}, COBL \cite{mizuta2024cobl}, and SafeDiffuser \cite{xiao2023safediffuser}.

\begin{algorithm}
\caption{\textbf{Guided Diffusion-Based Planning}}
\label{alg:1}
\begin{algorithmic}[1]
\Require Guidance energy $E$, diffusion model $\mu_{\theta}$, scaling factor $\alpha$, and covariance matrices $\Sigma$
\While{task is not completed}
    \State Observe the current state $s$
    \State Initialize a trajectory plan $\tau^N \sim \mathcal{N}(0, I)$ \Comment{Initialize a trajectory by sampling from a Gaussian distribution, serving as the starting point for the diffusion reverse process}
    \For{$i = N, \dots, 1$} 
        \State $\mu \gets \mu_{\theta}(\tau^i)$ \Comment{Diffusion reverse process: data-driven prior}
        \State $\tau^{i-1} \sim \mathcal{N}(\mu + \alpha \Sigma \nabla E(\mu),\; \Sigma)$ \Comment{Incorporate optimization objective guidance}
        \State $\tau^{i-1}_{s_0} \gets s$ \Comment{Enforce initial state constraint}
    \EndFor
    \State Execute the first action of the planned trajectory $\tau^0_{a_0}$
\EndWhile
\end{algorithmic}
\end{algorithm}

Notably, the update for the trajectory mean in step 6 of Algorithm \ref{alg:1} follows the same iterative update rule as MPPI. This suggests that guided diffusion-based planning effectively integrates both optimization objectives, as seen in MPPI and Policy Gradient, with data-driven distributions, as used in Diffusion models. Specifically, this approach alternates between:

\begin{enumerate}
    \item A step toward the data distribution, ensuring that sampled trajectories remain close to those seen in the dataset.
    \item A step toward optimizing an objective, by incorporating a guidance term that biases sampling toward trajectories that optimize the given cost or reward function.
\end{enumerate}

This iterative refinement—switching between data-driven priors and optimization-driven updates—suggests that guided diffusion provides a structured way to combine learning from demonstrations (as in Diffusion models) with control optimization (as in MPPI/Policy Gradient).

\bibliographystyle{plain}
\bibliography{main}

\end{document}